\documentclass[runningheads]{llncs}
\usepackage[T1]{fontenc}
\usepackage{graphicx}
\usepackage{hyperref}
\usepackage{algpseudocode}
\usepackage{algorithm}
\usepackage{float}
\usepackage{cite}
\usepackage{amsmath,amssymb,amsfonts}
\usepackage{textcomp}
\usepackage{xcolor}
\usepackage{hyperref}
\usepackage{todonotes}
\usepackage{booktabs}
\usepackage{breqn}
\usepackage{tikz}
\usetikzlibrary{shapes.geometric}

\def\BibTeX{{\rm B\kern-.05em{\sc i\kern-.025em b}\kern-.08em
   T\kern-.1667em\lower.7ex\hbox{E}\kern-.125emX}}

\begin{document}
\title{Fuzzy Representation of Norms}
%
%
\author{Ziba Assadi \and Paola Inverardi}

\authorrunning{Z. Assadi and P. Inverardi}

\institute{Gran Sasso Science Institute, L’Aquila, Italy}

\maketitle              
\begin{abstract}
Autonomous systems (AS) powered by AI components are increasingly integrated into the fabric of our daily lives and society, raising concerns about their ethical and social impact. To be considered trustworthy, AS must adhere to ethical principles and values. This has led to significant research on the identification and incorporation of ethical requirements in AS system design. A recent development in this area is the introduction of SLEEC (Social, Legal, Ethical, Empathetic, and Cultural) rules, which provide a comprehensive framework for representing ethical and other normative considerations. This paper proposes a logical representation of SLEEC rules and presents a methodology to embed these ethical requirements using test-score semantics and fuzzy logic. The use of fuzzy logic is motivated by the view of ethics as a domain of possibilities, which allows the resolution of ethical dilemmas that AI systems may encounter. The proposed approach is illustrated through a case study.

\end{abstract}
\section{Introduction}
Autonomous Systems (AS) are increasingly integrated into many aspects of daily life and society, raising growing concerns about their ethical and social implications\cite{Paola}. To be perceived as trustworthy, such systems must operate in accordance with well-defined ethical principles and human values. This requirement highlights the importance of embedding ethical considerations directly into their design and development processes.
A recent contribution in this area is the introduction of the SLEEC rules by Townsend et al.\cite{Townsend}, where SLEEC stands for social, legal, ethical, empathetic, and cultural normative rules. These categories represent high-level requirements that autonomous systems should not violate in their behaviour or decision-making.
In this paper, we propose an approach for translating ethical rules into a computational representation that can be embedded in autonomous systems, particularly robotic ones. We revisit the example of SLEEC rules for a healthcare robot introduced by Townsend et al. and present our own reformulation, which replaces the use of “unless” in the original model with explicit IF–THEN–ELSE structures. We argue that binary logic is insufficient to capture the nuances of human reasoning in ethically sensitive situations. To address this limitation, we extend boolean logic with fuzzy logic \cite{Zadeh2}, which enables the modeling of graded ethical reasoning and the representation of uncertainty in human decision-making.
Our approach builds on test-score semantics \cite{Zadeh_test_PRUF, Zadeh_test}, originally proposed by Zadeh (1982), to formalize vague or context-dependent concepts. By employing fuzzy logic, we allow certain ethical requirements to remain available at runtime as part of the system’s decision-making engine \cite{Paola2011}. This supports the system’s ability to handle, and in some cases resolve, ethical dilemmas the system may face  during interactions with humans.
Machine ethics has a long research history \cite{ACMsurvey}, including logical reasoning approaches such as deductive, non-monotonic, abductive, deontic, rule-based, event-calculus, knowledge-representation, and inductive logics \cite{RussellNorvig}. Building on these foundations, this work explores fuzzy logic as a method for handling uncertainty and supporting graded ethical reasoning in autonomous systems.
Existing approaches using fuzzy logic for ethical reasoning remain conceptual, descriptive or limited to ethical risk assessment (see Section \ref{sec:re}), leaving a gap in fully operational fuzzy ethical decision-making for autonomous systems. The approach proposed in this paper contributes filling this gap. 

The remainder of the paper is structured as follows. Section \ref{sec:2} reviews the existing formalization of SLEEC rules and their general structure. Section \ref{sec:3} discusses the nature of ethical rules and introduces fuzzy logic as an appropriate reasoning framework. Section \ref{sec:4} outlines our methodology and highlights the main contributions of this work. Section \ref{sec:5} applies the approach to the challenge of (soft) ethical dilemmas and illustrates a case study showing how fuzzy logic can support ethical decision-making. Finally, Section \ref{sec:conclusion} concludes the paper.

{\subsection{Related Works}\label{sec:re}
This sub-section shortly reviews existing work on fuzzy logic for machine ethics and positions our approach with respect to conceptual, descriptive, and partially implemented fuzzy ethical reasoning systems. 
Fuzzy logic has been widely used to represent ethical vagueness in AI systems, but most existing work remains conceptual, descriptive, or limited to ethical risk evaluation rather than full ethical decision-making. Conceptual approaches use fuzzy logic to bridge subjective values and objective data or to represent degrees of ethical conformity without implementing complete fuzzy inference (\cite{Inthorn2013MedicalFuzzy, Cervantes2016EthicalAgents, Kaufmann2022FuzzyEthizitat, Xu2025SemanticFuzzyEthics}), while survey work discusses fuzzy logic at a high level as one of several machine ethics paradigms (\cite{Narayanan2023MachineEthics}). More computational studies apply fuzzy reasoning to model ethical risks and moral justification—particularly in autonomous systems—without defuzzification or concrete decision outputs (\cite{Narayanan2019AVControlTakeover, Narayanan2019EthicalAVControl, DyoubLisi2024EthicalRisk, Dyoub2025ff4ERA}). Partial implementations exist, including simulation-based ethical reasoning in UAVs, fuzzy expert systems, and neuro-fuzzy or argumentation-based hybrids, but these focus on ethical risk assessment or representation rather than full fuzzy ethical reasoning pipelines (\cite{Smith2022EthicalUAV, Griffin2024FuzzyProlog, Sholla2021NeuroFuzzyIoT, Sholla2021FuzzyIoTethics, Baldi2025WeightedEthics}). Overall, fully implemented and validated fuzzy ethical decision-making systems for machine ethics remain unexplored.
}

\section{SLEEC Formalization and Refinement}\label{sec:2}
Our approach builds on the methodology proposed by Townsend et al. \cite{Townsend} for eliciting SLEEC requirements for autonomous systems. Further studies have addressed the operationalization of these rules, including conflict resolution and redundancy analysis \cite{feng2024analyzing,AAAI24,Yaman2025,Townsend2025}.
An example of a SLEEC rule, defined for a healthcare robot, is the following:

\noindent
  \textit{When the user tells the robot to open the curtains then the robot should open the curtains, {\textit unless} the user is `undressed' {\textit in which case} the robot does not open the curtains and tells the user `the curtains cannot be opened while you, the user, are undressed.'}   
    
\noindent
An additional defeater rule was also introduced, utilized here and in \cite{AAAI24}.

\noindent
\textit{... unless the user is `highly distressed' in which case the robot opens the curtains.}

\noindent
The formalizations in \cite{Townsend, AAAI24} employ the Quinean interpretation of {\textit unless} 
as an inclusive {\textit or} \cite{Quine}. However, using {\textit unless} as a functional connective introduces logical inconsistencies, due to the lack of equivalence between $p\vee q\vee r$ and $(p\vee q)\wedge(p\vee r)$ \cite{Ziba1}.
In \cite{Ziba2}, two linguistic interpretations of {\textit unless}—both functional and commutative—are presented. To avoid ambiguity, we follow the recommendation in \cite{Ziba1} and replace {\textit unless} with an explicit clearer and more suitable for machine interpretation IF–THEN–ELSE structure.
The reformulated rule is therefore:

\begin{table}[H]
\centering
\begin{tabular}{l}
\hline
\texttt{IF} the user is dressed \texttt{THEN} open the curtains, \\
\texttt{ELSE IF} the user is not highly distressed \texttt{THEN} do not open the curtains, \\
\texttt{ELSE} open the curtains. \\
\hline
\end{tabular}
\label{tab:curtain-rule}
\end{table}
\noindent
Then the general structure of a SLEEC rule -- {\textit When $c_0$ {\textit then} $a_0$ {\textit unless} $c_1$ {\textit in which case} $a_1$ {\textit unless} $c_2$ {\textit in which case} ...} -- according to the latter formalization for its conditions and actions (or defeaters) can be modeled as follows:
\begin{table}[H]
    \centering
    \renewcommand{\arraystretch}{0.8}  
    \setlength{\tabcolsep}{8pt}        
    \begin{tabular}{llll}
    \hline
       \texttt{if}       & $c_0$           & \texttt{then} & $a_0$ \\
       \texttt{else if}  & $c_1$           & \texttt{then} & $a_1$ \\
       \texttt{else if}  & $c_2$           & \texttt{then} & $a_2$ \\
       $\vdots$            & $\vdots$          & $\vdots$        & $\vdots$ \\
       \texttt{else if}  & $c_{n-1}$       & \texttt{then} & $a_{n-1}$ \\
       \texttt{else}\quad $a_n$ &  &  & \\
\hline
    \end{tabular}
    \label{tab:my_label}
\end{table}

\noindent
This structure ensures logical consistency and supports direct translation into computational models, facilitating the embedding of ethical rules into autonomous systems.

\section{The need for Possibility and Fuzzy Logic}\label{sec:3}
The inherent nature of ethical rules expressed in natural language is characterized by imprecision, gradability, and a tendency toward possibilities rather than probabilities. The degree of compatibility of a moral property can be more effectively represented through possibility. For instance, some moral properties—such as moral consistency, understood as the absence of contradictions in one’s ethical conduct—are conceptually feasible yet statistically rare. In such cases, their occurrence is possible but not necessarily probable. Therefore, for AS it is more meaningful to assess the possibility of an ethical rule being applicable rather than its likelihood.
Moreover, the uncertainty inherent in the interdependence of ethical rules often requires reasoning about the union of events. In this sense, evaluating the strength or applicability of ethical rules aligns more naturally with the logic of possibility than with probabilistic reasoning, defined for a union of elements $e_i$ as
$\operatorname{Possibility}\left(\bigcup_{i=1}^{n} e_i\right)=\bigvee_{i=1}^{n} \operatorname{Possibility}(e_i)=\max_{i=1}^{n} \operatorname{Possibility}(e_i)$  
and 
$\operatorname{Probability}\left(\bigcup_{i=1}^{n} e_i\right)=\sum_{k=1}^{n} (-1)^{k+1}\sum_{1 \le i_1 < \cdots < i_k \le n}\operatorname{Probability}\left(\bigcap_{j=1}^{k} e_{i_j}\right)$. 
Consider the following example related to user privacy and dressing preferences. Suppose a user specifies the requirement that they should not be undressed when the curtains are open. Let $\text{Poss(x)}$ denote the degree of possibility that the user is dressed in state x, and $\text{Prob(x)}$ denote the corresponding probability. 
If the user sets a threshold of $0.8$ to represent an acceptable level of being dressed, and expresses her comfort level with various clothing types as follows:
\begin{equation}
\label{eq:possibility-clothing}
\begin{array}{lll}
\text{Poss}(\text{Tops such as T-shirt, shirt, blouse, sweatshirt, etc.}) & = & 0.4 \\
\text{Poss}(\text{Bottoms such as skirt, pants, leggings, capri pants, etc.}) & = & 0.5 \\
\text{Poss}(\text{Dresses such as sundress, evening dress, gown, etc.}) & = & 1 \\
\text{Poss}(\text{Sleepwear such as nightgown, robe, etc.}) & = & 0.8 \\
\text{Poss}(\text{Accessories such as jewelry, sunglasses, watch, etc.}) & = & 0 \\
\text{Poss}(\text{Others such as socks, hat, belt, tie, etc.}) & = & 0.1
\end{array}
\end{equation}

\noindent
We can observe that reasoning with possibility aligns more closely with human perception than with probability. For instance, wearing socks may be more probable than wearing a sundress in winter, yet the sundress represents a complete state of being dressed according to the user’s ethical preference, whereas socks do not. Moreover, wearing multiple pairs of socks does not increase the degree of being dressed. Thus, the possibility of dressing is best represented by the maximum value among the relevant garment categories.
Probability and possibility can also diverge under changing circumstances, such as climate or context. For example, the probability of wearing socks increases in cold weather, while the possibility value assigned by the user remains constant. As a result, the inconsistency between probability and possibility \cite{Ziba3} becomes evident—reinforcing the argument for possibility-based ethical reasoning in adaptive systems.
Possibility theory \cite{Zadeh_possibility} provides a mathematical framework for handling imprecision through graded representations of uncertainty without relying on statistical information, making it particularly suitable for reasoning about ethical rules expressed in natural language.
Test-score semantics \cite{Zadeh_test_PRUF, Zadeh_test} complements this framework by assigning partial applicability scores to linguistic concepts, which are then aggregated to compute overall degrees of satisfaction.
Now, suppose the user requests that the robot open the curtains when she is highly distressed, even if the defined threshold of 0.8 is not met. Natural language expressions such as “highly distressed” are inherently vague and context-dependent. Mapping such linguistic terms into precise, machine-interpretable concepts requires a formalism capable of representing partial truth.
To achieve this, we apply fuzzy logic, which associates imprecise linguistic expressions with numerical values ranging between 0 and 1. Fuzzy logic extends classical boolean logic to handle the concept of partial truth, where truth values are not absolute but vary  between completely true and completely false. Given the non-absolute and graded nature of ethical reasoning, fuzzy logic provides an appropriate mechanism to represent and reason about ethical concepts that cannot be sharply defined.

\noindent
{{\bf Decision Algorithm on Dressing.} 
Let the user’s clothing preferences be $POSS(G)=\{Poss(g_1),\dots,Poss(g_n)\}$, the set of possibilities for garments $G=\{g_1,\dots,g_n\}$. The user’s threshold $T$ determines the dressing decision. The algorithm sums the distinct maximums (or union possibilities) of garment sets and compares the result with $T$ to decide on dressing. 
{At each iteration, the algorithm selects the maximum possibility value 
$\max_i Poss(g_i)$ (corresponding to the union operator $\vee$), accumulates distinct values in the set $D$, and outputs the decision function $F(D)\in\{0,1\}$, where $F(D)=1$ denotes dressed and $F(D)=0$ denotes undressed.}
\begin{algorithm}[H]
\caption{Check if User is Dressed and Compute $F(D)$}
\label{DAD}
\begin{algorithmic}[1]
\State $Sum \gets 0$, $D \gets \varnothing$
\While{$n \neq 0$}
    \State $\max \gets Poss\Big(\bigcup_{i=1}^{n} g_i\Big) \; (= \vee_{i=1}^{n} Poss(g_i) = \max_{i=1}^{n} Poss(g_i))$
    \If{$\max \geq T$} \State $F(D) \gets 1$; \textbf{exit} \Comment{USER IS DRESSED}
    \ElsIf{$\max \notin D$}
        \State $Sum \gets Sum + \max$
        \If{$Sum \geq T$} \State $F(D) \gets 1$; \textbf{exit} \Comment{USER IS DRESSED}
        \Else \State $D \gets D \cup \{\max\}$
        \EndIf
    \Else \State $POSS(G) \gets POSS(G) - \max$
    \EndIf
    \State $n \gets n - 1$
\EndWhile
\State $F(D) \gets 0$ \Comment{USER IS UNDRESSED}
\end{algorithmic}
\end{algorithm}

\section{Methodology}\label{sec:4}
Our semi-formal representation of SLEEC rules in natural language provides a foundation for their full logical formalization and subsequent embedding within AS. Achieving this transformation, however, requires the elimination of all forms of semantic ambiguity. Most natural language expressions inherently convey nuances of possibility and degree. For instance, in our example, the phrase ``highly distressed'' involves both the possibility of a user being ``distressed'' and the degree or intensity associated with the quantifier ``highly''. 
Our semantic framework is based on the possibility theory and the test-score semantics \cite{Zadeh_possibility, Zadeh_test_PRUF, Zadeh_test}, both introduced by Zadeh, which form the theoretical foundation of our method. {Test-score semantics computes partial scores of linguistic concepts under possibility-based reasoning. Partial scores are then aggregated to produce final degrees representing the satisfaction of each ethical rule.}
Our workflow is structured into three main stages. First, we identify an explanatory database (Descriptive data) composed of explicit and implicit necessities (Designation)—where the denotative and connotative aspects of natural language, as discussed in \cite{Rieger}, provide useful insights into the notions of explicitness and implicitness. Second, we perform possibility distribution and degree assignment to these necessities, followed by their combination and compatibility checking through fuzzification (Compilation). Finally, we quantify the resulting outcomes through defuzzification.

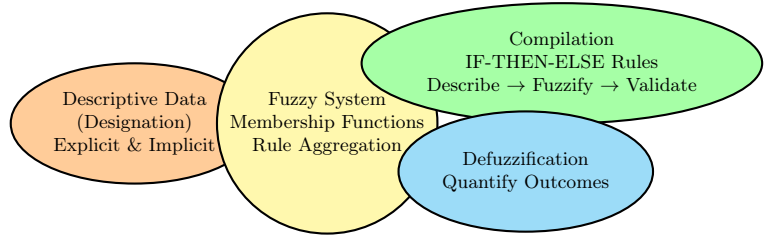
\begin{figure}[H]
\centering
\begin{tikzpicture}[scale=0.8, transform shape, font=\footnotesize, >=latex]

\node[
    draw,
    ellipse,
    thick,
    fill=orange!40,
    align=center,
    minimum width=2.6cm,
    minimum height=2cm
] (dd) at (1.5,0)
{Descriptive Data\\
(Designation)\\
Explicit \& Implicit};

\node[
    draw,
    circle,
    thick,
    fill=yellow!40,
    align=center,
    minimum width=2.2cm,
    minimum height=2.4cm
] (fs) at (4.7,0)
{Fuzzy System\\
Membership Functions\\
Rule Aggregation};

\node[
    draw,
    ellipse,
    thick,
    fill=green!35,
    align=center,
    minimum width=1.5cm,
    minimum height=2cm
] (comp) at (8.6,1)
{Compilation\\
IF-THEN-ELSE Rules\\
Describe $\rightarrow$ Fuzzify $\rightarrow$ Validate};

\node[
    draw,
    ellipse,
    thick,
    fill=cyan!35,
    align=center,
    minimum width=2.8cm,
    minimum height=2cm
] (df) at (8,-0.8)
{Defuzzification\\
Quantify Outcomes};


\end{tikzpicture}
\end{figure}

\noindent
This process embodies the foundation of test-score semantics, which associates every natural language concept with a degree of applicability. In our approach, concepts are partially evaluated through membership functions, their partial test-score degrees are aggregated to obtain overall scores, and compatibility among them is assessed using fuzzy rules.

\subsection{Preliminaries for Fuzziness}\label{sec:4.1}
Consider all components of SLEEC rules
as a  universe $U$ or domain of discourse.
The possibility function as a fuzzy membership function
$$\mathrm{POSSIBILITY} \triangleq \mu : U \rightarrow [0,1]$$
maps a variable of the universe to a value of the interval $[0,1]$.
Table \ref{mu} shows the possibility distribution or degrees of membership function for the distress variable $x\in\{\small{\text{Quite calm}},\cdots, \small{\text{Quite distressed}}\}$.
\begin{table}[H]
    \centering
    \begin{tabular}{c|c c c c c}
       $x$ & \small{Quite calm} & $\cdots$ & \small{Normal} & $\cdots$ & \small{Quite distressed} \\
       \hline
        $\mu(x)$ & $0$ & $\cdots$ & $0.5$ & $\cdots$ & $1$
    \end{tabular}
    \caption{Possibility distribution of distressed state}
    \label{mu}
\end{table}

\noindent
$F=\{(x_1,0),\cdots, (x_j, \mu_F(x_j)), \cdots, (x_n,1)\}$ 
is a representation of fuzzy sets as a collection of ordered pairs, each consisting of an element of the universe and its corresponding membership value.

\noindent
{\bf Designation.} We use Carnap's method to assign formal notation to entities according to their extension and intension nature \cite{Carnap}.
The term ``designator'' was introduced by Carnap for all expressions to which a semantic analysis is applied.
\begin{definition}
    The extension of a term or predicate is the corresponding class, and its intension is the corresponding property.
\end{definition}

\noindent
For instance, USER and CURTAINS are terms that are extensively designated according to their explicit meanings. In contrast, OPEN, DRESSED, and DISTRESSED are predicates that are intensively designated to capture their implicit senses—namely, OPEN referring to the action of grabbing and pulling the cord, DRESSED to the user having clothes on, and DISTRESSED to physiological or behavioral symptoms such as variations in blood pressure, body temperature, or heart rate, depending on age. The modifier HIGHLY functions as a quantification term, intensively designated to represent specific degrees of distress.

\noindent
{\bf Descriptive Data.} 
Descriptive data are derived for each SLEEC rule according to their extensional and intensional designations. In our conditional propositions, only the IF part requires a specific designation, as the THEN part corresponds to a boolean action. In the example under consideration, all designations are intensional and therefore implicit, and their descriptive data can be defined as
DD$\triangleq$DRESSED[Clothes;\ $\mu_{DC}$]+
DISTRESSED[Age;\ $\mu_A$, Blood Pressure;\ $\mu_{BP}$, Body Temperature;\ $\mu_{BT}$, Heart Rate;\ $\mu_{HR}$]+
HIGHLY[Distressed;\ $\mu_{HD}$].

\noindent
{\bf Compilation.} 
The formalized SLEEC rules, structured as nested IF-THEN-ELSE statements, exhibit the logical completeness required for system-level compilation and embedding. The process involves (i) describing the relevant data, (ii) fuzzifying non-absolute or graded concepts, and (iii) validating the resulting representation through compilation and compatibility testing.

\subsection{Fuzzification}\label{sec:4.2}
Fuzzification represents a controlled balance between crisp values and linguistic variables. It involves abstracting precise numerical data into vague or imprecise linguistic categories, thereby enabling the classification of large numeric ranges into a limited and interpretable set of linguistic labels.

\noindent
{\bf Dressed or Undressed.} 
Depending on personal preferences or cultural factors, users may define the concept of being dressed according to the number or combination of garments worn. In such cases, we propose the use of a discrete membership function to represent this variability. By applying discrete membership functions to upper- and lower-body garments and defining an appropriate threshold, the overall membership function for the concept of dressing can be constructed:
\begin{align}\label{DC}
    \mu_{DC}(x)=\begin{cases}
    \Sigma_i\mu_C(x_i)<T & 0 \\
    \Sigma_i\mu_C(x_i)\geq T & 1
\end{cases}
\end{align}
{where $\mu_C(x_i)=Poss(\bigcup_j x_{i_j})(=\max_j Poss(x_{i_j}))$.}

\noindent
For example, imagine a user having just one sock and a hat on. Assuming that
$\texttt{one sock}\triangleq x_{{i_s}},\ \texttt{hat}\triangleq x_{{i_h}}$ and 
$\mu_C(x_{{i_s}_j})=0.12,\ \mu_C(x_{{i_h}_j})=0.11,\ T=0.8$, 
first step of fuzzification results in 
$\Sigma_i\mu_C(x_i)=0.12+0.11=0.23<0.8=T$, 
Since the sum is less than the threshold, we proceed to the second phase. Here, we find that the user's membership function is defined as $\mu_{DC}(x)=0$. As a result, the system diagnoses the user as undressed due to the insufficient level of dressing indicated by the membership values.

\noindent
{\bf Distress Indicators.}  
Distress in individuals can stem from several physiological factors, including fluctuations in blood pressure, body temperature, and heart rate. These vital signs are classified based on age and we can divide them into three categories as Low, Medium, or High, depending on age group: Young, Middle, or Old.
For instance, let's consider a 40-year-old individual. According to scientific medical information provided by Harvard Health Publishing \footnote{https://www.health.harvard.edu/heart-health/what-your-heart-rate-is-telling-you}, the ranges for these indicators would typically be outlined in terms of what is considered normal, elevated, or concerning for that age group. This categorization helps healthcare professionals assess an individual’s health status and determine if they are experiencing distress due to abnormal readings in these vital signs.
\begin{align}\label{HR}
{\rm HR}_{40}(x)\triangleq\begin{cases}
      \texttt{Low} & x<60  \\
      \texttt{Low to Medium} & 60\leq x<90 \\
      \texttt{Medium} & 90\leq x\leq 153 \\
      \texttt{Medium to High} & 153<x\leq 180 \\
      \texttt{High} & x>180
  \end{cases}
\end{align}

\noindent
The speed of decision-making depends on different types of membership functions \cite{Korea}, such as Triangular, Trapezoidal, Piecewise linear, Gaussian and Singleton.
We use membership functions proposed by Zadeh \cite{Zadeh1-3, Zadeh-Q} for the possible distribution of age as Young, Middle-aged, and Old:

\footnotesize{
\begin{align}\label{A}
\begin{array}{l@{\qquad}l}
\begin{array}{l}
\mu_{A_Y}(x) =
\begin{cases}
1 & x \leq 25 \\
\cfrac{1}{1 + \left( \cfrac{x - 25}{5} \right)^2} & x > 25
\end{cases} \\[2ex]
\mu_{A_O}(x) =
\begin{cases}
0 & x \leq 50 \\
\cfrac{1}{1 + \left( \cfrac{x - 50}{5} \right)^{-2}} & x > 50
\end{cases}
\end{array}
&
\mu_{A_M}(x) =
\begin{cases}
0 & 0 < x < 35 \\
\cfrac{1}{1 + \left( \cfrac{x - 45}{4} \right)^4} & 35 \leq x < 45 \\
\cfrac{1}{1 + \left( \cfrac{x - 45}{5} \right)^2} & x \geq 45
\end{cases}
\end{array}
\end{align}
}

\noindent
And the trapezoidal membership function to convert the crisp values of the rest of the indicators to fuzzy sets \cite{Ziba1}:
\begin{equation}\label{eq:trapezoidal}
\mu(x;x_1,x_2,x_3,x_4)=\max(\min(\frac{x-x_1}{x_2-x_1},1,\frac{x_4-x}{x_4-x_3}),0)
\end{equation}

\noindent
{\bf Compatibility Test by Fuzzy Rules.} 
On the strength of the membership functions, the system recognizes a number in the interval $[0,1]$ as a degree for dressing and indicators of distress. Our fuzzy system requires a set of rules for aggregating the partially tested results into an overall assessment that reflects the compatibility of the SLEEC rule with the descriptive data. Essentially, these rules are needed to infer the user's state regarding dressing and distress before making a decision about opening the curtains.
Fuzzy rules enable the system to make decisions based on imprecision, as they convert fuzzy sets into linguistic values. In the context of the formalized SLEEC rule applied in nursing homes: 

\begin{table}[H]
    \centering
    \begin{tabular}{llll}
       \texttt{if}  & $c_0$ & \texttt{then} & $a_0$ \\
       \texttt{else if}  & $c_1$ & \texttt{then} & $a_1 \equiv\neg a_0$ \\
       \texttt{else}\quad $a_2\equiv a_0$ &  &  &
    \end{tabular}
    \label{tab:my_label}
\end{table}

\noindent
Proposition $c_0$, which refers to a dressed user, can be categorized as boolean due to its boolean membership function. Similarly, proposition $a_0$ is also boolean, as it relates to the action of opening or not opening the curtains. Proposition $c_1$, which concerns a user being not highly distressed, remains somewhat ambiguous at this stage. The linguistic variables combined with logical connective symbols are essential for constructing if-then rules. These fuzzy if-then rules are pivotal in controlling the output variables. The inference engine selects the optimal variables, emulating boolean logic with basic operators. This variable indicates the user's level of distress, taking into account measurements such as age, blood pressure, body temperature, and heart rate.
Table \ref{FR0} summarizes the fuzzy rule base for the nursing home SLEEC system, consisting of up to \(3^5 = 243\) rules. The \(i\)-th rule combines age (\(A_i\)), blood pressure (\(BP_i\)), heart rate (\(HR_i\)), and body temperature (\(BT_i\)) to infer a distress level (\(D_i\)) using the linguistic terms shown in the table. 
As an example: \textbf{IF} \(A\) is \emph{Old} \(\wedge\) \(BP\) is \emph{High} \(\wedge\) \(HR\) is \emph{High} \(\wedge\) \(BT\) is \emph{High}, \textbf{THEN} \(D\) is \emph{High}.

\begin{table}[H]
\centering
\begin{tabular}{ccccc}
\footnotesize 
Rule & \textbf{\texttt{IF}}  & {\scriptsize Linguistic Terms} & \textbf{\texttt{THEN}}  & {\scriptsize Linguistic Terms} \\
\hline
$R_1$ & $\boldsymbol{A_i}\boldsymbol\wedge\boldsymbol{BP_i}\boldsymbol\wedge\boldsymbol{HR_i} \boldsymbol\wedge\boldsymbol{BT_i}$ & 
{\scriptsize
$\begin{cases}
A_i \in 
\left\{
\begin{array}{l}
\text{Young} \\
\text{Middle-aged} \\
\text{Old}
\end{array}
\right\} \\
BP_i, HR_i, BT_i \in 
\left\{
\begin{array}{l}
\text{Low} \\
\text{Medium} \\
\text{High}
\end{array}
\right\}
\end{cases}$} &
$\boldsymbol{D_i}$ & 
{\scriptsize
$D_i \in 
\left\{
\begin{array}{l}
\text{Low} \\
\text{Medium} \\
\text{High}
\end{array}
\right\}$} \\
\hline
$\vdots$ & $\vdots$ &  $\vdots$ & $\vdots$ &  $\vdots$ \\
\hline
$R_{243}$ & $\boldsymbol{A_i}\boldsymbol\wedge\boldsymbol{BP_i}\boldsymbol\wedge\boldsymbol{HR_i} \boldsymbol\wedge\boldsymbol{BT_i}$ & 
{\scriptsize
$\begin{cases}
A_i \in 
\left\{
\begin{array}{l}
\text{Young} \\
\text{Middle-aged} \\
\text{Old}
\end{array}
\right\} \\
BP_i, HR_i, BT_i \in 
\left\{
\begin{array}{l}
\text{Low} \\
\text{Medium} \\
\text{High}
\end{array}
\right\}
\end{cases}$} & 
$\boldsymbol{D_i}$ & 
{\scriptsize
$D_i \in 
\left\{
\begin{array}{l}
\text{Low} \\
\text{Medium} \\
\text{High}
\end{array}
\right\}$} \\
\hline
\end{tabular}
\caption{Fuzzy rules for the nursing home SLEEC system ($3^5 = 243$ possible rules).}
\label{FR0}
\end{table}

\subsection{Defuzzification}\label{sec:4.3}
Inferring fuzzy rules leaves us with fuzzy outputs, which should be converted to numeric and crisp values during the defuzzification process. We employ the center of gravity (COG) defuzzification method:
\begin{equation}
\label{eq:defuzzification}
D^*(\mathbf{x}) = 
\frac{\sum_i D_i \cdot \min(\mu_{A_i}(x_1), \mu_{BP_i}(x_2), \mu_{HR_i}(x_3), \mu_{BT_i}(x_4))}
{\sum_i \min(\mu_{A_i}(x_1), \mu_{BP_i}(x_2), \mu_{HR_i}(x_3), \mu_{BT_i}(x_4))},
\end{equation}
\noindent
where $\mathbf{x} = (x_1, x_2, x_3, x_4)$ are the crisp input values for Age, Blood Pressure, 
Heart Rate, and Body Temperature. 
Each membership function 
$\mu_{A_i}(x_1)$, $\mu_{BP_i}(x_2)$, $\mu_{HR_i}(x_3)$, $\mu_{BT_i}(x_4)$ measures how strongly 
the input belongs to the corresponding fuzzy set. This method computes a crisp value by averaging 
the typical values of all rules, where each rule contributes according to how strongly it is satisfied 
by the inputs. In our formula, this contribution is measured by
$w_i = \min\big(\mu_{A_i}(x_1), \mu_{BP_i}(x_2), \mu_{HR_i}(x_3), \mu_{BT_i}(x_4)\big)$.
Intuitively, rules that are more strongly satisfied by the given inputs (corresponding to a higher $w_i$ value)
pull the final result toward their typical output, so the resulting $D^*$ naturally reflects the dominant rules
without needing to know the fuzzy expressions in detail:
$D^* = \frac{\sum_i D_i \cdot w_i}{\sum_i w_i}$.

\noindent
Algorithm \ref{Dstar} summarizes the fuzzy inference procedure—fuzzification, rule evaluation, aggregation, and defuzzification—for identifying highly distressed users. The crisp inputs—Age ($a$), Blood Pressure ($bp$), Heart Rate ($hr$), and Body Temperature ($bt$)—are first converted into fuzzy values using the corresponding membership functions (labels: L = Low, M = Medium, H = High; O = Old, Y = Young, M = Middle). Each fuzzy rule contributes to the overall distress assessment according to how well the inputs match the rule conditions, and the final distress score is obtained through the center-of-gravity defuzzification method.

\begin{algorithm}[H]
\caption{Distress Assessment with Fuzzy Rules}
\label{Dstar}
\begin{algorithmic}[1]
\State Compute membership degrees:
\[
\begin{aligned}
&v_{A_Y} \gets \mu_{A_Y}(a), \quad v_{A_M} \gets \mu_{A_M}(a), \quad v_{A_O} \gets \mu_{A_O}(a)\\
&v_{BP_L} \gets \mu_{BP_L}(bp), \; v_{BP_M} \gets \mu_{BP_M}(bp), \; v_{BP_H} \gets \mu_{BP_H}(bp)\\
&v_{HR_L} \gets \mu_{HR_L}(hr), \; v_{HR_M} \gets \mu_{HR_M}(hr), \; v_{HR_H} \gets \mu_{HR_H}(hr)\\
&v_{BT_L} \gets \mu_{BT_L}(bt), \; v_{BT_M} \gets \mu_{BT_M}(bt), \; v_{BT_H} \gets \mu_{BT_H}(bt)
\end{aligned}
\]

\State Initialize: $numerator \gets 0$, $denominator \gets 0$, $i \gets 0$

\For{(A, $v_A$) in $\{(A_Y,v_{A_Y}), (A_M,v_{A_M}), (A_O,v_{A_O})\}$}
  \For{(BP, $v_{BP}$) in $\{(BP_L,v_{BP_L}), (BP_M,v_{BP_M}), (BP_H,v_{BP_H})\}$}
    \For{(HR, $v_{HR}$) in $\{(HR_L,v_{HR_L}), (HR_M,v_{HR_M}), (HR_H,v_{HR_H})\}$}
      \For{(BT, $v_{BT}$) in $\{(BT_L,v_{BT_L}), (BT_M,v_{BT_M}), (BT_H,v_{BT_H})\}$}
        \State $i \gets i + 1$ \Comment{Define rule $R_i$}
        
        \State Determine rule output $D_i$ based on the fuzzy rule table:  
        \[
        D_i \in \{\text{Low}=0.2,\ \text{Medium}=0.5,\ \text{High}=0.8\}
        \]
        (select the corresponding level according to $(A,BP,HR,BT)$ combination)

        \State $w_i \gets \min(v_A, v_{BP}, v_{HR}, v_{BT})$
        \State Accumulate for defuzzification:
        \[
        numerator \gets numerator + D_i \cdot w_i
        \]
        \[
        denominator \gets denominator + w_i
        \]
      \EndFor
    \EndFor
  \EndFor
\EndFor

\State Compute defuzzified distress:
\[
D^\ast(a, bp, hr, bt) \gets \frac{numerator}{denominator}
\]
\end{algorithmic}
\end{algorithm}

\section{Application}\label{sec:5}
The interaction between autonomous systems and humans can be enhanced by incorporating both subjective and objective requirements. Our approach establishes a framework for automated decision-making by assigning truth values to subjective as well as objective characteristics. 
{In the following two subsections, we first demonstrate the implementation of our method through a concrete case study based on the SLEEC rule—introduced and formalized in Section~\ref{sec:2}—to evaluate dressing and distress, and subsequently discuss how this formulation can be used to resolve ethical dilemmas in human–robot interaction.}

{\subsection{Automated recognition of user characteristics}
In Section \ref{sec:2}, we introduced and formalized an example of SLEEC rules, and in Section \ref{sec:4}, we presented a step-by-step explanation of our method based on this example. Here, we employ the formalized example as a case study to implement the algorithm and clarify the method.
\textit{\small When the robot is asked to open the curtains:
If the user is dressed
then open the curtains, 
else if the user is not highly distressed
then do not open the curtains, 
else open the curtains. 
}
We go through the steps of our procedure using this example:

\noindent
1. Designation. 
\noindent
Since ``user'' and ``curtains'' are extensively designated words, they are stored in the library. 
Even if ``open (the curtains)'' is intentional, the robot is already able to learn it in advance by asking the user, and then store it in the library. 

\noindent
2. Descriptive Data. 
\noindent
Our descriptive data regarding the conditions is as follows (results are boolean):

\noindent
DD$\triangleq$DRESSED[Clothes;\ $\mu_{DC}$]+
DISTRESSED[Age;\ $\mu_A$, Blood Pressure;\ $\mu_{BP}$, Body Temperature;\ $\mu_{BT}$, Heart Rate;\ $\mu_{HR}$]+
HIGHLY[Distressed;\ $\mu_{HD}$].

\noindent
3. Fuzzification. 
\noindent
In this step, we first define the membership functions of the descriptive data, $\mu_{DC}$ (see (\ref{DC})) and $\mu_A$ (see (\ref{A})), and, in particular, the membership functions for blood pressure, body temperature, and heart rate, $\mu_{BP}$, $\mu_{BT}$, and $\mu_{HR}$ (see (\ref{eq:trapezoidal})), as previously described in Section~\ref{sec:4.2}. Subsequently, we establish fuzzy rules of the form:

\begin{table}[htbp]
\centering
\small
\setlength{\tabcolsep}{8pt} 
\begin{tabular}{@{} l l @{}} 
\toprule
Fuzzy rule & 
$R_i:\; \text{IF }\ \mu_{A_i}\wedge\mu_{BP_i}\wedge\mu_{HR_i}\wedge\mu_{BT_i}\ \text{THEN }\ \mu_{D_i},\quad i=1,\ldots,n$ \\[4pt]
\midrule
Membership sets &
$\displaystyle
\begin{aligned}
\mu_{A_i}&\in\{\mu_{A_Y},\mu_{A_M},\mu_{A_O}\},\\[-2pt]
\mu_{BP_i}&\in\{\mu_{BP_L},\mu_{BP_M},\mu_{BP_H}\},\\[-2pt]
\mu_{HR_i}&\in\{\mu_{HR_L},\mu_{HR_M},\mu_{HR_H}\},\\[-2pt]
\mu_{BT_i}&\in\{\mu_{BT_L},\mu_{BT_M},\mu_{BT_H}\},\\[-2pt]
\mu_{D_i}&\in\{\mu_{D_L},\mu_{D_M},\mu_{D_H}\},\quad \mu_{D_i}(x)\in[0,1]
\end{aligned}$ \\
\bottomrule
\end{tabular}
\caption{Compact representation of fuzzy rules and associated membership-function sets.}
\label{tab:fuzzy-compact}
\end{table}

\noindent
4. Defuzzification and Decision-making. 
\noindent
The robot determines its actions based on the measured $D^\ast$ value (\ref{eq:defuzzification}).

\[D^\ast(a, bp, hr, bt)=
\frac{\Sigma_{i=1}^{n} D_i\cdot\big(\min(\mu_{A_i}(a),\ \mu_{BP_i}(bp),\ \mu_{HR_i}(hr),\ \mu_{BT_i}(bt))\big)}{\Sigma_{i=1}^{n}\min(\mu_{A_i}(a),\ \mu_{BP_i}(bp),\ \mu_{HR_i}(hr),\ \mu_{BT_i}(bt))}\]

\noindent
Given the complexity and length of the numerical computation required for defuzzification, this work presents only the methodological procedure and resulting outcome, while omitting the step-by-step arithmetic for clarity. 
Figure \ref{D_F_S} illustrates the defuzzification of the aggregated fuzzy outputs for distress.
Each colored region represents one of the distress levels—Low, Medium, or High. The height of each region is adjusted based on how well the input data (age, blood pressure, body temperature, and heart rate) satisfy the conditions of that rule, which is calculated as the minimum of their membership values ($\min\big(\mu_{A_i}(x_A), \mu_{BP_i}(x_{BP}), \mu_{BT_i}(x_{BT}), \mu_{HR_i}(x_{HR})$). 
$D_i$ is the centroid (center) of the trapezoidal distress output for rule $i$, representing the typical numeric value of that distress level (in Figure \ref{D_F_S}: Low $\sim 0.2$, Medium $\sim 0.5$, and High $\sim 0.8$ distress).

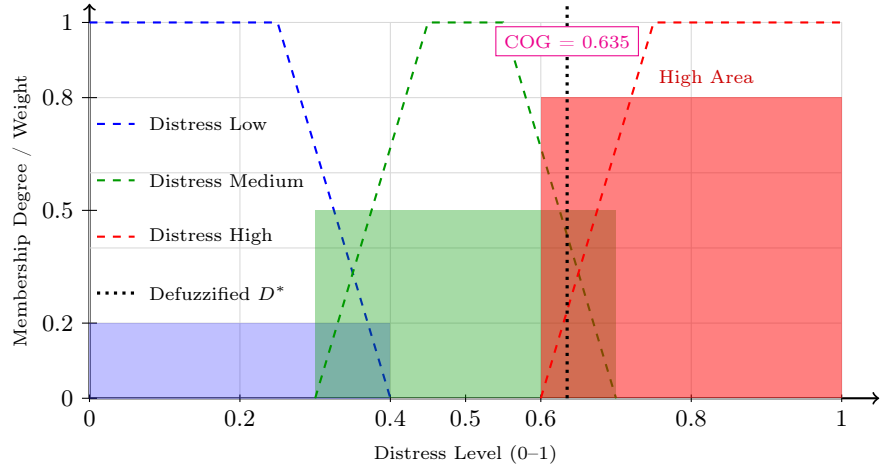
\begin{figure}[H]
\centering
\begin{tikzpicture}[xscale=10, yscale=5]

\draw[->, thick] (0,0) -- (1.05,0); 
\draw[->, thick] (0,0) -- (0,1.05); 

\node[rotate=90, font=\scriptsize] at (-0.09,0.5) {Membership Degree / Weight};
\node[font=\scriptsize] at (0.5,-0.15) {Distress Level (0--1)};

\foreach \x in {0,0.2,...,1}
    \draw[gray!30] (\x,0) -- (\x,1);
\foreach \y in {0,0.2,...,1}
    \draw[gray!30] (0,\y) -- (1,\y);

\foreach \x/\label in {0/0, 0.2/0.2, 0.4/0.4, 0.5/0.5, 0.6/0.6, 0.8/0.8, 1/1}
    \draw (\x,0.01) -- (\x,-0.01) node[below, font=\footnotesize] {\label};

\foreach \y/\label in {0/0, 0.2/0.2, 0.5/0.5, 0.8/0.8, 1/1}
    \draw (0.01,\y) -- (-0.01,\y) node[left, font=\footnotesize] {\label};

\draw[blue, dashed, thick] (0,1) -- (0.25,1) -- (0.4,0);
\fill[blue, opacity=0.25] (0,0) rectangle (0.4,0.2);

\draw[green!60!black, dashed, thick] (0.3,0) -- (0.45,1) -- (0.55,1) -- (0.7,0);
\fill[green!60!black, opacity=0.35] (0.3,0) rectangle (0.7,0.5);

\draw[red, dashed, thick] (0.6,0) -- (0.75,1) -- (1,1);
\fill[red, opacity=0.5] (0.6,0) rectangle (1,0.8);

\draw[black, dotted, very thick] (0.635,0) -- (0.635,1.05);
\node[draw=magenta, text=magenta, fill=white, font=\scriptsize] at (0.635,0.95) {COG = 0.635};

\node[red!80!black, font=\scriptsize] at (0.82,0.85) {High Area};

\begin{scope}[shift={(0.01,0.73)}, scale=0.25]
    \draw[blue, dashed, thick] (0,0) -- (0.2,0);
    \node[right, font=\scriptsize] at (0.25,0) {\!Distress Low};

    \draw[green!60!black, dashed, thick] (0,-0.6) -- (0.2,-0.6);
    \node[right, font=\scriptsize] at (0.25,-0.6) {\!Distress Medium};

    \draw[red, dashed, thick] (0,-1.2) -- (0.2,-1.2);
    \node[right, font=\scriptsize] at (0.25,-1.2) {\!Distress High};

    \draw[black, dotted, very thick] (0,-1.8) -- (0.2,-1.8);
    \node[right, font=\scriptsize] at (0.25,-1.8) {\!Defuzzified \(D^\ast\)};
\end{scope}

\end{tikzpicture}
\caption{\footnotesize Defuzzification of the aggregated fuzzy distress output.}

\label{D_F_S}
\end{figure}

\noindent
When two fuzzy sets overlap (e.g., Medium and High Distress), 
the region with the larger aggregated area pulls the centroid (defuzzified value, $D^\ast$) toward itself. 
As a result, $D^\ast$ tends to be located within the region contributing more to the overall fuzzy area. 
In this example, a defuzzified value of $D^\ast = 0.635$ lies within the overlap between the Medium and High regions 
but is pulled toward the High side, indicating a high distress state. 
{In this fuzzy configuration, when the defuzzified value lies in the overlap between Medium and High Distress,
the dominant distress level is determined by the larger aggregated area; values above 0.6 are dominated by
the High distress region and are therefore interpreted as High Distress.}
Therefore, the threshold of distress for triggering the curtain-opening action should be set to $0.6$. 
Algorithm~\ref{DCO} then determines whether to open the curtain based on the user’s dressing status 
(from Algorithm~\ref{DAD}) and the defuzzified distress level (from Algorithm~\ref{Dstar}).

\begin{algorithm}[H]
\caption{Dynamic Curtain Opener (Fuzzy Distress)}
\label{DCO}
\begin{algorithmic}[1]
\State Run Algorithm~\ref{DAD} \Comment{Determine if user is dressed}
\State $v_{DC} \gets F(D)$  \Comment{1 if dressed, 0 if not}
\State Run Algorithm~\ref{Dstar} \Comment{Compute fuzzy distress $D^\ast \in [0,1]$}

\If{$v_{DC} = 1$} 
    \State \textbf{Open curtains} \Comment{User is dressed}
\ElsIf{$v_{DC} = 0 \wedge D^\ast \le 0.6$} 
    \State \textbf{Do not open curtains} \Comment{User not dressed and not highly distressed}
\Else
    \State \textbf{Open curtains} \Comment{User not dressed but highly distressed}
\EndIf
\end{algorithmic}
\end{algorithm}

\subsection{Overcoming Ethical Dilemmas}
Our method provides a means of addressing ethical dilemmas in human–robot interaction scenarios. The scenario to which we applied our fuzzy approach above poses a clear ethical dilemma: opening the curtains may violate the user’s privacy, whereas refraining from action may conflict with the user’s expressed preferences and potentially compromise their health in the presence of high distress.
To address this dilemma, our approach estimates the user’s level of distress using fuzzy inference and dynamically balances competing ethical considerations, enabling the robot to decide whether preserving privacy or acting in support of the user’s autonomy and health is ethically justified.

\section{Conclusions}\label{sec:conclusion}
In our work, we embed SLEEC rules into the AI system by defining a dataset that appropriately represents each rule’s conceptual dimensions and by generating corresponding distributions using relevant membership functions.
As expected, some rules require the introduction of new concepts—such as our definition of “highly distressed”. This process results in a preliminary dataset but does not yet yield executable commands. Inspired by previously formalized IF–THEN SLEEC rules \cite{Ziba1}, we employ fuzzy rules to perform aggregation and compatibility testing. At this stage, the presence of non-boolean linguistic values necessitates their defuzzification into single numeric values, making them interpretable and actionable within the system.
Finally, we address ethical dilemmas concerning the preservation of human autonomy and life by incorporating fuzzy logic into the management of user-defined health directives. {Inspired by prior work in healthcare robotics \cite{Feng},}this approach aims to identify the subtle boundary where the respect for human dignity intersects with the imperative to preserve life in robotic decision-making.

\bibliographystyle{splncs04}
\bibliography{base}

@String{Computing = "Computing" }

@String{Computer = "{IEEE} Computer" }

@String{Springer = "Springer-Verlag" }

@inproceedings{Ziba1,
  author    = {Ziba Assadi},
  title     = {Logical Formalisms for Ethics},
  booktitle = {GoodIT'24: Proceedings of the 2024 International Conference on Information Technology for Social Good},
  year      = {2024},
  pages     = {416--419},
  doi       = {10.1145/3677525.3678691}
}

@inproceedings{Ziba2,
  author    = {Ziba Assadi},
  title     = {Non-Quineian unless},
  booktitle = {XXI Brazilian Logic Conference},
  year      = {2025},
  pages     = {263},
  url       = {https://ssrn.com/abstract=5236844}
}

@inproceedings{Ziba3,
  author    = {Ziba Assadi and Paola Inverardi},
  title     = {Fuzzy Logic in Ethical AI},
  booktitle = {The 9th Women in Logic Workshop},
  year      = {2025},
  pages     = {27--29},
  url       = {https://dx.doi.org/10.2139/ssrn.5241896}
}

@book{Carnap,
  author    = {Rudolf Carnap},
  title     = {Meaning and necessity: A study in semantics and modal logic},
  publisher = {University of Chicago Press},
  year      = {1988},
  edition   = {30th}
}

@inproceedings{feng2024analyzing,
  author    = {Nick Feng and Lina Marsso and Sinem Getir Yaman and others},
  title     = {Analyzing and debugging normative requirements via satisfiability checking},
  booktitle = {ICSE '24: Proceedings of the IEEE/ACM 46th International Conference on Software Engineering},
  year      = {2024},
  article   = {214},
  pages     = {1--12},
  doi       = {10.1145/3597503.3639093}
}

@misc{Feng,
  author = {Nick Feng and Lina Marsso},
  title  = {Supplementary material for: Analyzing and debugging normative requirements via satisfiability checking},
  year   = {2024},
  url    = {https://github.com/NickF0211/LEGOS-SLEEC}
}

@incollection{Paola,
  author    = {Paola Inverardi},
  title     = {The Challenge of Human Dignity in the Era of Autonomous Systems},
  booktitle = {Perspectives on Digital Humanism},
  editor    = {H. Werthner and E. Prem and E.A. Lee and C. Ghezzi},
  publisher = {Springer},
  year      = {2022},
  pages     = {25--29},
  doi       = {10.1007/978-3-030-86144-5_4}
}

@inproceedings{Paola2011,
  author    = {Paola Inverardi and Marco Mori},
  title     = {Requirements models at run-time to support consistent system evolutions},
  booktitle = {2011 2nd International Workshop on Requirements@Run.Time},
  year      = {2011},
  pages     = {1--8},
  publisher = {IEEE Computer Society},
  doi       = {10.1109/RERUNTIME.2011.6046241}
}

@book{Quine,
  author    = {Willard Van Orman Quine},
  title     = {Methods of Logic},
  publisher = {New York},
  year      = {1959},
  edition   = {Revised Ed}
}

@article{Rieger,
  author  = {Burghard B Rieger},
  title   = {Feasible Fuzzy Semantics. On some problems of how to handle word meaning empirically},
  journal = {Words, Worlds, and Contexts. New Approaches in Word Semantics (Research in Text Theory)},
  year    = {1981},
  volume  = {6},
  pages   = {193--209}
}

@book{RussellNorvig,
  author    = {Stuart J. Russell and Peter Norvig},
  title     = {Artificial Intelligence: A Modern Approach},
  publisher = {Pearson Education Limited},
  year      = {2016},
  address   = {Kuala Lumpur, Malaysia}
}

@article{Korea,
  author  = {Suntae Kim and Malrey Lee and Jongho Lee},
  title   = {A study of fuzzy membership functions for dependence decision-making in security robot system},
  journal = {Neural Computing \& Application},
  year    = {2017},
  volume  = {28},
  number  = {1},
  pages   = {155--164},
  doi     = {10.1007/s00521-015-2044-3}
}

@article{Townsend,
  author  = {Beverley Townsend and Colin Paterson and T. T. Arvind and others},
  title   = {From Pluralistic Normative Principles to Autonomous-Agent Rules},
  journal = {Minds and Machines},
  year    = {2022},
  volume  = {32},
  number  = {4},
  pages   = {683--715},
  doi     = {10.1007/s11023-022-09614-w}
}

@article{Townsend2025,
  author  = {Beverley Townsend and Katherine J. Parnell and Sinem Getir Yaman and Gabriel Nemirovsky and Radu Calinescu},
  title   = {Normative conflict resolution through human–autonomous agent interaction},
  journal = {Journal of Responsible Technology},
  year    = {2025},
  volume  = {21},
  pages   = {100114},
  doi     = {10.1016/j.jrt.2025.100114}
}

@article{ACMsurvey,
  author  = {Suzanne Tolmeijer and Markus Kneer and Cristina Sarasua and others},
  title   = {Implementations in machine ethics: A survey},
  journal = {ACM Computing Surveys},
  year    = {2021},
  volume  = {53},
  number  = {6},
  pages   = {1--38},
  doi     = {10.1145/3419633}
}

@article{AAAI24,
  author  = {Nicolas Troquard and Martina De Sanctis and Paola Inverardi and Patrizio Pelliccione and Gian Luca Scoccia},
  title   = {Social, Legal, Ethical, Empathetic, and Cultural Rules: Compilation and Reasoning},
  journal = {Proceedings of the AAAI Conference on Artificial Intelligence},
  year    = {2024},
  volume  = {38},
  number  = {20},
  pages   = {22385--22392},
  doi     = {10.1609/aaai.v38i20.30245}
}

@article{Yaman2025,
  author  = {Sila G. Yaman and Pedro Ribeiro and Ana Cavalcanti and others},
  title   = {Specification, validation and verification of social, legal, ethical, empathetic and cultural requirements for autonomous agents},
  journal = {Journal of Systems and Software},
  year    = {2025},
  volume  = {220},
  pages   = {112229},
  doi     = {10.1016/j.jss.2024.112229}
}

@article{Zadeh-Q,
  author  = {Lotfi A. Zadeh},
  title   = {Quantitative fuzzy semantics},
  journal = {Information Sciences},
  year    = {1971},
  volume  = {3},
  number  = {2},
  pages   = {159--176},
  doi     = {10.1016/S0020-0255(71)80004-X}
}

@article{Zadeh1-3,
  author  = {Lotfi A. Zadeh},
  title   = {The concept of a linguistic variable and its application to approximate reasoning—I, II, III},
  journal = {Information Sciences},
  year    = {1975},
  pages   = {199--249, 301--357, 43--80},
  doi     = {10.1016/0020-0255(75)90036-5, 10.1016/0020-0255(75)90046-8, 10.1016/0020-0255(75)90017-1}
}

@article{Zadeh_possibility,
  author  = {Lotfi A. Zadeh},
  title   = {Fuzzy sets as a basis for a theory of possibility},
  journal = {Fuzzy Sets and Systems},
  year    = {1978},
  volume  = {1},
  number  = {1},
  pages   = {3--28},
  doi     = {10.1016/0165-0114(78)90029-5}
}

@article{Zadeh_test,
  author  = {Lotfi A. Zadeh},
  title   = {Test-score semantics for natural languages},
  journal = {COLING '82: Proceedings of the 9th Conference on Computational Linguistics},
  year    = {1982},
  volume  = {1},
  pages   = {425--430},
  doi     = {10.3115/991813.991881}
}

@article{Zadeh_test_PRUF,
  author  = {Lotfi A. Zadeh},
  title   = {Test-score semantics for natural languages and meaning-representation via PRUF},
  journal = {Fuzzy Sets, Fuzzy Logic, and Fuzzy Systems},
  year    = {1996},
  pages   = {542--586},
  doi     = {10.1142/9789814261302\_0026}
}

@article{Zadeh2,
  author  = {Lotfi A. Zadeh},
  title   = {Is there a need for fuzzy logic?},
  journal = {Information Sciences},
  year    = {2008},
  volume  = {178},
  number  = {13},
  pages   = {2751--2779},
  doi     = {10.1016/j.ins.2008.02.012}
}

@incollection{Inthorn2013MedicalFuzzy,
  author    = {Inthorn, Julia},
  title     = {Medical Ethics, Fuzzy Logic and Shared Decision Making},
  booktitle = {Fuzziness and Medicine: Philosophical Reflections and Application Systems in Health Care},
  pages     = {85--95},
  year      = {2013},
  doi       = {10.1007/978-3-642-36527-0_5},
}

@article{Cervantes2016EthicalAgents,
  author    = {Cervantes, José-Antonio and Rodríguez, Luis F. and López, Sergio and Ramos, Francisco and Robles, Francisco},
  title     = {Autonomous Agents and Ethical Decision-Making},
  journal   = {Cognitive Computation},
  volume    = {8},
  number    = {2},
  pages     = {278--296},
  year      = {2016},
  doi       = {10.1007/s12559-015-9362-8}
}

@article{Narayanan2019AVControlTakeover,
  author    = {Narayanan, Ajit},
  title     = {When is it Right and Good for an Intelligent Autonomous Vehicle to Take Over Control (and Hand it Back)?},
  journal   = {arXiv preprint arXiv:1901.08221},
  year      = {2019},
  note      = {Available at: \url{https://arXiv.org/abs/1901.08221}}
}

@article{Sholla2021NeuroFuzzyIoT,
  author    = {Sholla, Sahil and Mir, Roohie Naaz and Chishti, Mohammad Ahsan},
  title     = {A Neuro Fuzzy System for Incorporating Ethics in the Internet of Things},
  journal   = {Journal of Ambient Intelligence and Humanized Computing},
  volume    = {12},
  number    = {1},
  pages     = {1487--1501},
  year      = {2021},
  doi       = {10.1007/s12652-020-02217-2},
}

@article{Kaufmann2022FuzzyEthizitat,
  author    = {Kaufmann, Michael and Meier, Andreas},
  title     = {Fuzzy Ethizität: Radar für ethische Künstliche Intelligenz},
  journal   = {HMD Praxis der Wirtschaftsinformatik},
  volume    = {59},
  number    = {2},
  pages     = {538--555},
  year      = {2022},
  doi       = {10.1365/s40702-022-00857-w}
}

@mastersthesis{Smith2022EthicalUAV,
  author    = {Smith, Gavin Giovanni},
  title     = {Design of Ethical Autonomous Agents for Unmanned Aerial Vehicles using Fuzzy Logic},
  school    = {Florida Institute of Technology},
  year      = {2022},
  note      = {Available at: \url{https://repository.fit.edu/etd/934/}}
}

@article{Narayanan2023MachineEthics,
  author    = {Narayanan, Ajit},
  title     = {Machine Ethics and Cognitive Robotics},
  journal   = {Current Robotics Reports},
  volume    = {4},
  number    = {2},
  pages     = {33--41},
  year      = {2023},
  doi       = {10.1007/s43154-023-00098-9},
}

@inproceedings{DyoubLisi2024EthicalRisk,
  author    = {Dyoub, Abeer and Lisi, Francesca Alessandra},
  title     = {Towards Ethical Risk Assessment of Symbiotic AI Systems with Fuzzy Rules},
  booktitle = {CEUR Workshop Proceedings},
  volume    = {3881},
  pages     = {36--49},
  year      = {2024},
  note      = {Available at: \url{https://ceur-ws.org/Vol-3881/paper5.pdf}}
}

@article{Dyoub2025ff4ERA,
  author    = {Dyoub, Abeer and Letteri, Ivan and Lisi, Francesca A.},
  title     = {ff4ERA: A New Fuzzy Framework for Ethical Risk Assessment in AI},
  journal   = {arXiv preprint arXiv:2508.00899},
  year      = {2025},
  note      = {Available at: \url{https://arXiv.org/abs/2508.00899}}
}

@misc{Xu2025SemanticFuzzyEthics,
  author    = {Xu, Jingyuan},
  title     = {Semantic Representation of Fuzzy Ethical Boundaries in AI},
  year      = {2025},
  doi       = {10.5772/intechopen.1012203},
  note      = {Available at: \url{https://www.researchgate.net/publication/394958155_Semantic_Representation_of_Fuzzy_Ethical_Boundaries_in_AI}}
}

@inproceedings{Baldi2025WeightedEthics,
  author    = {Baldi, Paolo and D’Asaro, Fabio Aurelio and Dyoub, Abeer and Lisi, Francesca A.},
  title     = {Weighted Assumption Based Argumentation to Reason About Ethical Principles and Actions},
  booktitle = {CEUR Workshop Proceedings},
  year      = {2025},
  doi       = {10.48550/arXiv.2506.18056},
}

@inproceedings{Narayanan2019EthicalAVControl,
  author    = {Narayanan, Ajit},
  title     = {Ethical Judgement in Intelligent Control Systems for Autonomous Vehicles},
  booktitle = {2019 Australian \& New Zealand Control Conference (ANZCC)},
  pages     = {231--236},
  year      = {2019},
  doi       = {10.1109/ANZCC47194.2019.8945790},
}

@article{Sholla2021FuzzyIoTethics,
  author   = {Sholla, Sahil and Mir, Roohie Naaz and Chishti, Mohammad Ahsan},
  title    = {A Fuzzy Logic-Based Method for Incorporating Ethics in the Internet of Things},
  journal  = {International Journal of Ambient Computing and Intelligence},
  year     = {2021},
  volume   = {12},
  number   = {3},
  pages    = {98--122},
  doi      = {10.4018/IJACI.2021070105},
  note     = {Available at: \url{https://ums.iust.ac.in/UploadedDocuments/EmployeeDocuments/ResearchPapers/1965_R0I9UY.pdf}}
}

@inproceedings{Griffin2024FuzzyProlog,
  author    = {Griffin, Harriet and Ghahremani, Mani and Gegov, Alexander},
  title     = {A Fuzzy Expert System Based Extension of SWI-Prolog for Evaluating AI Ethics},
  booktitle = {2024 IEEE 12th International Conference on Intelligent Systems (IS)},
  pages     = {1--6},
  year      = {2024},
  doi       = {10.1109/IS61756.2024.10705249},
}

\end{document}